\documentclass[runningheads]{llncs}
\usepackage{graphicx}
\usepackage{tikz}
\usepackage{comment}
\usepackage{amsmath,amssymb} % define this before the line numbering.
\usepackage{color}
\usepackage{xspace}

\makeatletter
\newbox\cvprrulerbox
\newcount\cvprrulercount
\newcount\cvprrulercounttmp
\newdimen\cvprruleroffset
\newdimen\cv@lineheight
\newdimen\cv@boxheight
\newbox\cv@tmpbox
\newcount\cv@refno
\newcount\cv@tot
\DeclareRobustCommand\onedot{\futurelet\@let@token\@onedot}
\def\@onedot{\ifx\@let@token.\else.\null\fi\xspace}
\def\etal{\emph{et al}\onedot}
\def\eg{\emph{e.g}\onedot}

\newcommand \footnoteonlytext[1]
{
    \let \mybackup \thefootnote
    \let \thefootnote \relax
    \footnotetext{#1}
    \let \thefootnote \mybackup
    \let \mybackup \imareallyundefinedcommand
}

\usepackage[accsupp]{axessibility}  % Improves PDF readability for those with disabilities.

\begin{document}
% \linenumbers
\pagestyle{headings}
\mainmatter
\def\ECCVSubNumber{100}  % Insert your submission number here

\title{Stripformer: Strip Transformer for Fast Image Deblurring} % Replace with your title

\begin{comment}
\titlerunning{ECCV-22 submission ID \ECCVSubNumber} 
\authorrunning{ECCV-22 submission ID \ECCVSubNumber} 
\author{Anonymous ECCV submission}
\institute{Paper ID \ECCVSubNumber}
\end{comment}
%******************

%\begin{comment}
\titlerunning{Stripformer: Strip Transformer for Fast Image Deblurring}

\author{Fu-Jen Tsai\inst{1}\index{Tsai, Fu-Jen}\and
Yan-Tsung Peng\inst{2}\index{Peng, Yan-Tsung}\and
Yen-Yu Lin\inst{3}\index{Lin, Yen-Yu}\and
Chung-Chi Tsai\index{Tsai, Chung-Chi}\inst{4}\and
Chia-Wen Lin\index{Lin, Chia-Wen}\inst{1}
}
\authorrunning{F.-J. Tsai et al.}
\institute{National Tsing Hua University, Taiwan \\
\email{cwlin@ee.nthu.edu.tw, fjtsai@gapp.nthu.edu.tw}\and
National Chengchi University, Taiwan\\
\email{ytpeng@cs.nccu.edu.tw}
\and
National Yang Ming Chiao Tung University, Taiwan\\
\email{lin@cs.nycu.edu.tw}\and
Qualcomm Technologies, Inc., San Diego\\
\email{chuntsai@qti.qualcomm.com}
}
%\end{comment}
%******************

\maketitle

\begin{abstract}
Images taken in dynamic scenes may contain unwanted motion blur, which significantly degrades visual quality. Such blur causes short- and long-range region-specific smoothing artifacts that are often directional and non-uniform, which is difficult to be removed. Inspired by the current success of transformers on computer vision and image processing tasks, we develop, Stripformer, a transformer-based architecture that constructs intra- and inter-strip tokens to reweight image features in the horizontal and vertical directions to catch blurred patterns with different orientations. It stacks interlaced intra-strip and inter-strip attention layers to reveal blur magnitudes. 
In addition to detecting region-specific blurred patterns of various orientations and magnitudes, Stripformer is also a token-efficient and parameter-efficient transformer model, demanding much less memory usage and computation cost than the vanilla transformer but works better without relying on tremendous training data. Experimental results show that Stripformer performs favorably against state-of-the-art models in dynamic scene deblurring.
\end{abstract}
\section{Introduction}

Blur coming from object movement or camera shaking causes a smudge in taken images, often unwanted for photographers and affecting the performance of subsequent computer vision applications. Dynamic scene image deblurring aims to recover sharpness from a single blurred image, which is difficult since such blur is usually globally and locally non-uniform, and only limited information can be utilized from the single image. 

Conventional approaches usually exploit prior knowledge for single image deblurring due to its ill-posedness. Some methods simplify this task by assuming that only uniform blur exists~\cite{Cho_2009_ACM,Fergus2006}; However, it is often not the case for real-world dynamic scenes. Some works utilize prior assumptions to remove non-uniform blur~\cite{5206802,6909767,Pan_2016_cvpr}.
However, non-uniform blur is usually region-specific, which is hard to be modeled by the specific priors, often making these works fail. In addition, these methods typically involve solving a non-convex optimization problem, leading to high computation time.

\begin{figure}[t!]
\centering
\includegraphics[width=0.8\columnwidth]{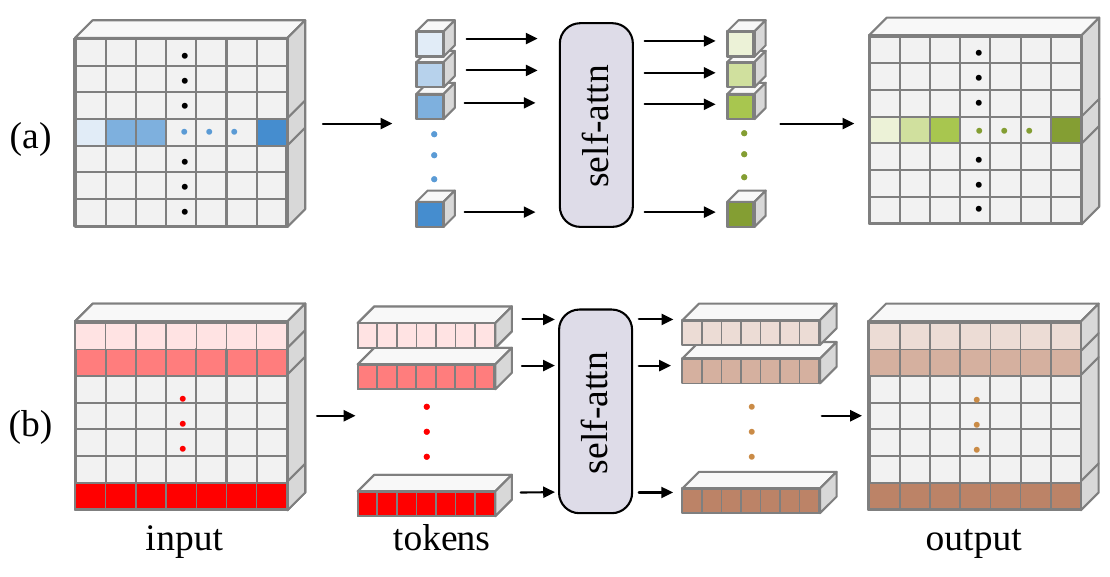}
\caption{
(a) Horizontal intra-strip attention (Intra-SA-H) encodes pixel dependence within the same horizontal strip.
Vertical intra-strip attention (Intra-SA-V) is symmetrically constructed.
(b) Horizontal inter-strip attention (Inter-SA-H) captures strip-wise correlations.
Inter-SA-V is similarly established for vertical strips.
Horizontal and vertical intra-strip and inter-strip attention works jointly to explore blur orientations.
Stacking interlaced intra-strip and inter-strip attention layers reveals blur magnitudes. 
}
\label{fig:introduction}
\end{figure}

Deblurring has made significant progress using deep learning. Based on convolutional neural networks (CNNs), several studies have improved the deblurring performance based on recurrent architectures, such as multi-scale (MS) \cite{gao2019dynamic,Nah_2017_CVPR}, multi-patch (MP) \cite{SAPN2020,Zhang_2019_CVPR}, and multi-temporal (MT) \cite{MT_2020_ECCV} recurrent architectures. However, blur from dynamic scenes, in general, is non-uniform and regionally directional, requiring a better model design to explore global and local correlations from blur in one image. 

Recently, motivated by the success of transformers~\cite{vaswani2017attention} which exploit attention mechanisms for natural language processing, researchers have explored transformer-based architectures to address computer vision tasks and obtained promising results, such as image classification~\cite{dosovitskiy2020vit}, object detection~\cite{DETR} and low-level vision~\cite{IPT}. We explore the self-attention mechanisms used in transformers to deal with blurred patterns with different magnitudes and orientations.

Transformer architectures can be generally classified into two categories: a pure encoder-decoder architecture~\cite{vaswani2017attention} and a hybrid architecture~\cite{DETR}. The former treats image patches of a fixed size $n \times n$ as tokens and takes these tokens as input. To preserve fine-grained information, the number of parameters grows proportional to $n^2$, like IPT~\cite{IPT}, which requires numerous parameters and relies on a large amount of training data (over $1M$ images) to achieve competitive results. The hybrid architecture extracts embedding features of an input image using additional models such as CNNs before the transformer is applied. The architecture demands high memory and computation consumption due to its pixel-wise attention, up to $\mathcal{O}(H^2W^2)$ for an input image or feature maps of resolution $H\times W$. A trade-off between compactness and efficiency is present.

In transformers, similar tokens mutually attend to each other. The attention mechanism can capture all-range information, which is essential to superior performance in image deblurring but tends to have a large memory requirement. In addition, since blur patterns are often region-specific and hard to catch by a deblurring model, we instead try to specify each region-specific pattern using its orientation and magnitude. Thus, it suffices to attain orientation and magnitude information at each position for deblurring.

We leverage these observations to address the trade-off between a pure transformer and a hybrid transformer. In turn, we propose a token-efficient and parameter-efficient hybrid transformer architecture, called Stripformer, exploiting both intra-strip and inter-strip attention, shown in Figure~\ref{fig:introduction}, to reassemble the attended blur features. 
The intra-strip tokens, forming the intra-strip attention, carry local pixel-wise blur features. In contrast, the inter-strip tokens, forming the inter-strip attention, bear global region-wise blur information.

The designs of intra-strip and inter-strip attention are inspired by~\cite{Jian_2015_CVPR}, which projects blur motions into horizontal and vertical directions in the Cartesian coordinate system for estimating the motion blur field of a blurred image.
The intra- and inter-strip attention contains horizontal and vertical branches to capture blur patterns. 
The captured horizontal and vertical features stored at each pixel offer sufficient information for the subsequent layer to infer the blur pattern orientation at that pixel.
Moreover, sequential local-feature extraction by successive intra-strip blocks obtains multi-scale features, which reveal blur pattern magnitudes.
It turns out that we stack multi-head intra-strip and inter-strip attention blocks to decompose dynamic blur into different orientations and magnitudes, and can remove short- and long-range blurred artifacts from the input image.

The intra- and inter-strip attention in our Stripformer is developed based on the inductive bias of image deblurring.
It also results in an efficient transformer model.
Since the intra-strip and inter-strip tokens are fewer than those used in the vanilla attention, Stripformer requires much less memory and computation costs than the vanilla transformer. 
Therefore, Stripformer works better without relying on tremendous training data. Extensive experimental results show that Stripformer performs favorably against state-of-the-art (SOTA) deblurring models in recovered image quality, memory usage, and computational efficiency.
The source code is available at {\color{blue}\url{https://github.com/pp00704831/Stripformer}}.

\section{Related Work}
\noindent\textbf{Deblurring via CNN-based Architectures.}
Single image deblurring using CNN-based architectures has achieved promising performance. Most of these successful architectures are recurrent and can be roughly classified into three types: Multi-scale (MS), multi-patch (MP), and multi-temporal (MT) models. Nah \etal~\cite{Nah_2017_CVPR} propose an MS network by a {\em coarse-to-fine} strategy to restore a sharp image on different resolutions gradually. Zhang~\etal \cite{Zhang_2019_CVPR} utilize an MP method by building a hierarchical deblurring model. Motivated by MS and MP, Park~\etal \cite{MT_2020_ECCV} propose an MT deblurring model via incremental temporal training in the original spatial scale to preserve more high-frequency information for reliable deblurring. In addition, Kupyn~\etal \cite{Kupyn_2019_ICCV} suggest using conditional generative adversarial CNN networks to restore high-quality visual results.

\noindent\textbf{Attention Mechanism.}
Attention mechanisms~\cite{vaswani2017attention} have been commonly used in the fields of image processing~\cite{parmar2018image,SAGAN_2019_PMLR} and computer vision~\cite{hu2018senet,wang2018non} to encode long-range dependency in the extracted features. Specifically to deblurring, attention mechanisms can help learn cross-pixel correlations to better address non-uniform blur~\cite{RADN_2020_ECCV,SAPN2020}. Hence, transformers with multi-head self-attention to explore local and global correlations would be a good choice for deblurring. 

Hou~\etal \cite{hou2020strip} exploit horizontal and vertical one-pixel long kernels to pool images and extract context information for scene parsing, called strip pooling, initially designed for contextual information extraction instead of deblurring. We extend strip tokens to intra- and inter-strip attentions for better capturing blurred patterns.
CCNet \cite{huang2019ccnet} utilizes the criss-cross attention to capture the global image dependencies for semantic segmentation, working a bit similar to the proposed intra-strip attention. The criss-cross attention computes pixel correlations horizontally and vertically in a joint manner. In contrast, we construct horizontal and vertical intra-strips separately and calculate their intra-strip attentions parallelly. Moreover, intra-strip attention works together with the region-wise inter-strip attention to capture blurred patterns locally and globally.

\noindent\textbf{Vision Transformer.}
Unlike conventional CNN architectures, the transformers are originally proposed for natural language processing (NLP), utilizing multi-head self-attention to model global token-to-token relationships. 
Recently, transformers have achieved comparable or even better performance than CNN models in several vision applications such as image classification~\cite{dosovitskiy2020vit}, object detection~\cite{DETR}, semantic segmentation~\cite{Ranftl21}, inpainting~\cite{yan2020sttn}, and super-resolution~\cite{yang2020learning}.
% Fu-Jen
Take Vision Transformers (ViT)~\cite{dosovitskiy2020vit} for image classification as an example. 
ViT generates tokens for the Multi-head Self-Attention (MSA) from the pixels or patches of an image. 
The former flattens the three-dimensional feature maps $X \in \mathbb{R}^{H \times W \times C}$ produced by a CNN model to a two-dimensional tensor of size $\mathbb{R}^{HW \times C}$. Its global self-attention mechanism requires up to $\mathcal{O}(H^2W^2)$ space complexity for each head of MSA, which is memory-demanding.
The latter uses patches instead of pixels as tokens, like~\cite{IPT}, where each token is a patch of size $8 \times 8$. However, it needs lots of parameters ($114M$ used in \cite{IPT}) to preserve all the channel dimensions and keep the spatial information. 
Moreover, transformers with more parameters rely on more training data for stable optimization. 
In~\cite{IPT}, the model requires to be pre-trained on ImageNet with more than one million annotated images for deraining, denoising, and super-resolution to obtain competitive results.

To address the issue of high memory consumption of transformers, Liu~\etal propose Swin~\cite{liu2021Swin}, a transformer architecture that uses a sliding window to make it token-efficient and realize local attention. However, Swin~\cite{liu2021Swin} does not consider high-resolution global attention, which is crucial to some dense prediction tasks.
For example, images with dynamic scene blur commonly have local and global blur artifacts, requiring deblurring models to consider short-range and long-range pixel correlations. 
Chu~\etal~\cite{chu2021Twins} proposed Twins, which utilizes locally-grouped self-attention (LSA) by local window attention and global sub-sampled attention (GSA) by pooling key and value features to the size $7\times7$ for classification. However, it is not suitable for high-resolution dense prediction tasks such as image deblurring by only using $7\times7$ features.
In our design, we simultaneously leverage the prior observation of blurred patterns to reduce the number of tokens and parameters.

The proposed Stripformer is a token-efficient transformer with its space complexity of only $\mathcal{O}(HW(H+W))$ and $\mathcal{O}(H^2 + W^2)$ for intra- and inter-strip attention, respectively, where $H$ and $W$ are the height and width of the input image. It is much less than the vanilla transformer's $\mathcal{O}(H^2W^2)$. In addition, our model uses much fewer parameters ($20M$) than IPT~\cite{IPT} ($114M$), thus not needing a large amount of training data to achieve even better performance.

\begin{figure}[t]
    \begin{center}
    \includegraphics[width=1\textwidth]{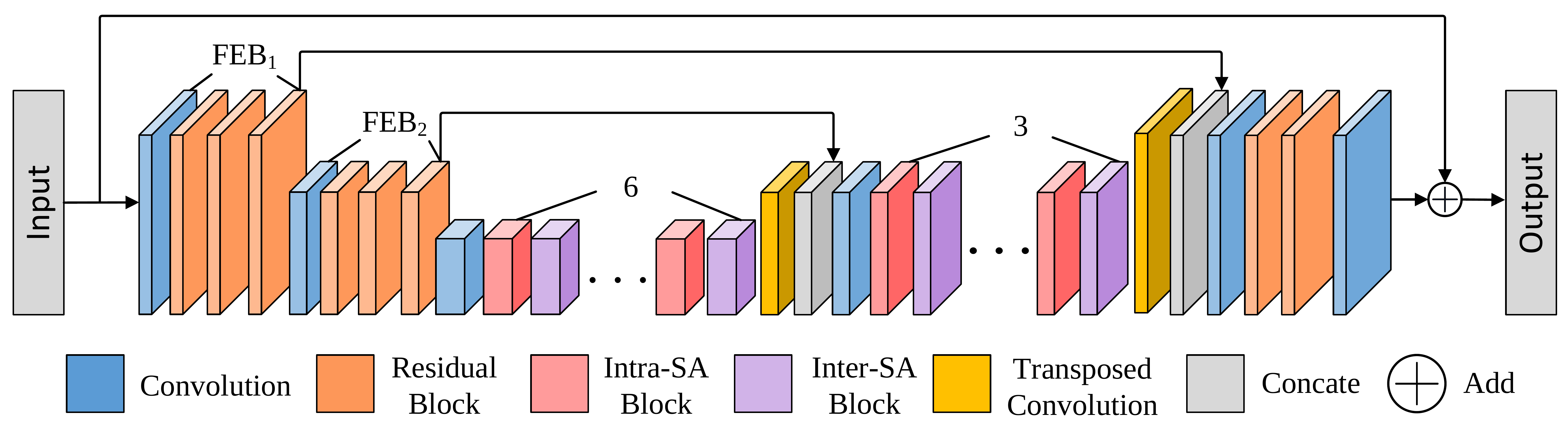}
    \end{center}
    \caption{Architecture of Stripformer. We utilize shallow convolution embedding with intra- and inter-strip attention blocks for image deblurring.}
    \label{fig:architecture}
\end{figure}

\section{Proposed Method}
Images captured in dynamic scenes often suffer from blurring, where the blur artifacts could have various orientations and magnitudes. 
The proposed Stripformer is a transformer-based architecture that leverages intra- and inter-strip tokens to extract blurred patterns with different orientations and magnitudes. The intra- and inter-strip tokens contain horizontal and vertical strip-wise features to form multi-head intra-strip attention and inter-strip attention blocks to break down region-specific blur patterns into different orientations. Through their attention mechanisms, intra-strip and inter-strip features can be reweighted to fit short- and long-range blur magnitudes.     

\begin{figure}[t]
    \begin{center}
    \includegraphics[width=1\textwidth]{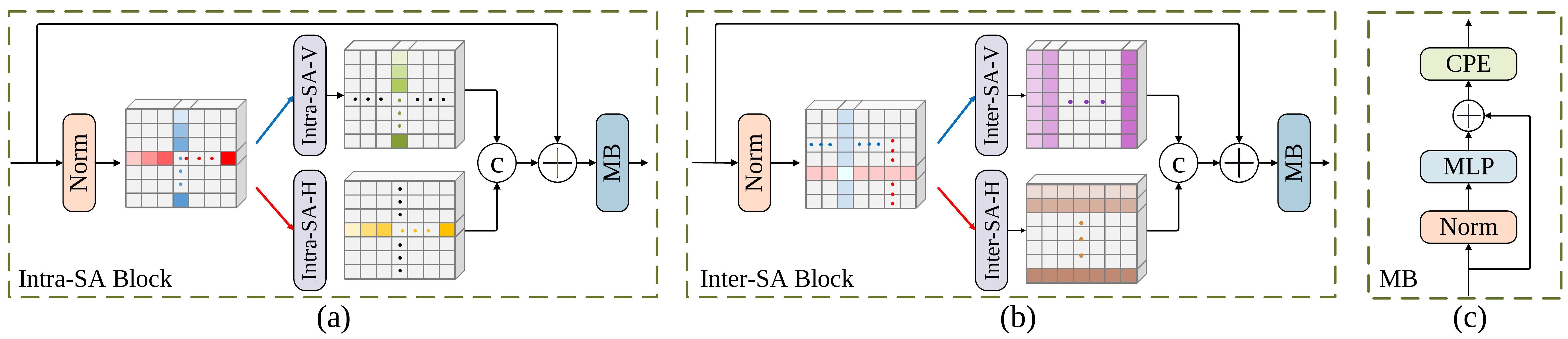}
    \end{center}
    \caption{Illustration of (a) Intra-Strip Attention (Intra-SA) Block, (b) Inter-Strip Attention (Inter-SA) Block, where \textcircled{c} denotes concatenation, and (c) MLP Block (MB), where CPE denotes the conditional position encoding~\cite{chu2021conditional}.} 
    \label{fig:intra_and_inter}
\end{figure}

Figure~\ref{fig:architecture} demonstrates the model design of Stripformer, which is a residual encoder-decoder architecture starting with two Feature Embedding Blocks (FEBs) to generate embedding features. Since a FEB downsamples the input, the output resolution is one-fourth of the input after two FEBs. 
Next, it stacks a convolution layer with interlaced Intra-SA and Inter-SA blocks on the smallest and second-smallest scales. As shown in Figure~\ref{fig:intra_and_inter}, in Intra-SA and Inter-SA blocks, we perform horizontal and vertical intra-strip or inter-strip attention to produce multi-range strip-shaped features to catch blur with different magnitudes and orientations. We adopt transposed convolution for upsampling. Its output features are concatenated with those generated from the encoder on the same scale.
Lastly, Stripformer ends with two residual blocks and a convolution layer with a residual connection to the input blurred image. In the following, we detail the main functional modules: FEBs, Intra-SA blocks, and Inter-SA blocks. 

\subsection{Feature Embedding Block (FEB)}\label{subsec:FEB}
For a vanilla transformer, the input image is usually divided into patches before feeding them to a transformer~\cite{IPT,dosovitskiy2020vit}, meaning that features in each patch are flattened to yield a token. However, this could cause spatial pixel correlations to be lost due to flattened pixels and require numerous parameters because of its self-attention mechanism. Instead, we use two FEBs, each of which consists of one convolutional layer and three residual blocks to generate feature embedding without losing spatial information.  
\subsection{Intra-SA and Inter-SA Blocks}\label{subsec:SWT}
The core of Stripformer is Intra-SA and Inter-SA blocks. We detail their designs as follows.

\noindent\textbf{Intra-SA Block.}
As shown in Figure~\ref{fig:intra_and_inter} (a), an Intra-SA block consists of two paralleled branches: horizontal intra-strip attention (Intra-SA-H) and vertical intra-strip attention (Intra-SA-V). Let the input features of an intra-strip block be ${X} \in \mathbb{R}^{H \times W \times C}$, where $H$, $W$, and $C$ represent the height, the width, and the number of channels, respectively. We first process them with a LayerNorm layer (\textbf{Norm}) followed by a $1 \times 1$ convolution layer (\textbf{Conv}) with $C$ filters to obtain the input features, described as 
\begin{equation}
    (X^h, X^v) = \textbf{Conv}(\textbf{Norm}(X)),\\
    \label{eqn:input}
\end{equation}
where $X^h$ and $X^v \in \mathbb{R}^{H \times W \times D}$ stand for the input features for Intra-SA-H and Intra-SA-V, respectively, where $D=\frac{C}{2}$.

For the horizontal intra-strip attention, we split the input features $X^h$ into $H$ non-overlapping horizontal strip $X^h_{i} \in \mathbb{R}^{W \times D},~i=\{1, 2, ..., H\}$. Each strip $X^h_{i}$ has $W$ tokens with $D$ dimensions. Next, we generate queries, keys, and values associated with $X^h_{i}$ as $Q^{h}_{ij}$, $K^{h}_{ij}$, and $V^{h}_{ij} \in \mathbb{R}^{W \times \frac{D}{m}}$ for the multi-head attention mechanism as 
\begin{equation}
    \begin{gathered} 
    (Q^{h}_{ij}, K^{h}_{ij}, V^{h}_{ij}) = (X^{h}_{i}{P}_j^{Q}, X^{h}_{i}{P}_j^{K}, X^{h}_{i}{P}_j^{V}), 
    \end{gathered}
    \label{eqn:qkv}
\end{equation}
where ${P}_j^{Q}$, ${P}_j^{K}$, and ${P}_j^{V}\in \mathbb{R}^{D\times \frac{D}{m}},~j \in\{1, ..., m\}$, representing linear projection matrices for the query, key, and value with the multi-head attention. Here, we set the number of heads to five, $m=5$. 
The multi-head attended feature $O^{h}_{ij} \in \mathbb{R}^{W \times \frac{D}{m}}$ for one horizontal strip is calculated as
\begin{equation}
    O^{h}_{ij} = \mathbf{Softmax}(\frac{Q^{h}_{ij}(K^{h}_{ij})^{T}}{\sqrt{D/m}})V^{h}_{ij},
\end{equation}
whose space complexity is $\mathcal{O}(W^{2})$. 
We concatenate the multi-head horizontal features $O^{h}_{ij} \in \mathbb{R}^{W \times \frac{D}{m}}$ along the channel dimension to generate $O^{h}_{i} \in \mathbb{R}^{W \times D}$ and fold all of them into three-dimensional tensors $O^{h} \in \mathbb{R}^{H \times W \times D}$ as the Intra-SA-H output. Symmetrically, the vertical intra-strip attention produces the multi-head attended feature for one vertical strip, denoted as $O^{v}_{ij} \in \mathbb{R}^{H \times \frac{D}{m}}$, whose space complexity is $\mathcal{O}(H^{2})$. After folding all the vertical features, the Intra-SA-V output denotes as $O^{v} \in \mathbb{R}^{H \times W \times D}$.

We then concatenate them to feed into a $1 \times 1$ convolution layer with a residual connection to the original input features ${X}$ to obtain the attended features $O_{attn} \in \mathbb{R}^{H \times W \times C}$ as
\begin{equation}
    \begin{gathered} 
    O_{attn} = \mathbf{Conv}(\mathbf{Concate}(O^{h}, O^{v})) + X.
    \end{gathered}
    \label{eqn:attn}
\end{equation}

An MLP block, as illustrated in Figure~\ref{fig:intra_and_inter} (c), is then applied to $O_{attn}$. Specifically, we use LayerNorm, feed-forward MultiLayer Perceptron ($\mathbf{MLP}$) with a residual connection, and the Conditional Positional Encodings~\cite{chu2021conditional} ($\mathbf{CPE}$), a $3\times3$ depth-wise convolution layer with a residual connection, to generate the final output $O_{intra} \in R^{H \times W \times C}$ as
\begin{equation}
    \begin{gathered} 
    O_{intra} = \mathbf{CPE}(\mathbf{MLP}(\mathbf{Norm}(O_{attn})) + O_{attn}).
    \end{gathered}
    \label{eqn:mlp}
\end{equation}
The total space complexity of Intra-SA is $\mathcal{O}(HW^2+WH^2)$ for $H$ horizontal and $W$ vertical strips.

\noindent\textbf{Inter-SA Block.}
As shown in Figure~\ref{fig:intra_and_inter} (b), an Inter-SA block also consists of two paralleled branches: horizontal inter-strip attention (Inter-SA-H) and vertical inter-strip attention (Inter-SA-V). Inter-SA is a strip-wise attention that regards each strip feature as a token. 
We process the input like Intra-SA using Eq.~(\ref{eqn:input}) to generate input features $X^h$ and $X^v \in \mathbb{R}^{H \times W \times D}$ for Inter-SA-H and Inter-SA-V, respectively.

For the horizontal inter-strip attention, we generate the multi-head queries, keys, and values by linear projection matrices as Eq.~(\ref{eqn:qkv}), where we abused the notation for simplicity as $Q_j^{h}$, $K_j^{h}$, and $V_j^{h} \in \mathbb{R}^{H \times W \times \frac{D}{m}}$. 
Next, we reshape $Q_j^{h}$, $K_j^{h}$, and $V_j^{h}$ to two-dimensional tensors with the size of ${H\times \frac{D^{h}}{m}}$, where $D^{h} = W\times D$, representing $H$ horizontal strip tokens with the size of $\frac{D^{h}}{m}$. Then, the output features $O_j^{h} \in \mathbb{R}^{H\times \frac{D^{h}}{m}}$ is calculated as
\begin{equation}
    %A^{h} = Softmax(\frac{Q^{h}(K^{h})^{T}}{\sqrt{D^{h}}}), \\
    O_j^{h} = \mathbf{Softmax}(\frac{Q_j^{h}(K_j^{h})^{T}}{\sqrt{D^{h}/m}})V_j^{h},
\end{equation}
whose space complexity is $\mathcal{O}(H^{2})$.
Symmetrically, the vertical inter-strip attention generates the multi-head attended features $O_j^{v} \in \mathbb{R}^{W\times \frac{D^{v}}{m}}$, where $D^{v} = H\times D$. Its space complexity is $\mathcal{O}(W^{2})$ in the attention mechanism.

Lastly, we concatenate the multi-head horizontal and vertical features along the channel dimension to be $O^{h} \in \mathbb{R}^{H\times D^{h}}$ and $O^{v} \in \mathbb{R}^{W\times D^{v}}$ and reshape them back to three-dimensional tensors with the size of ${H \times W \times D}$. Similar to Intra-SA in Eq.~(\ref{eqn:attn}) and Eq.~(\ref{eqn:mlp}), we can generate the final output $O_{inter} \in R^{H \times W \times C}$ for an Inter-SA block. The total space complexity of Inter-SA is $\mathcal{O}(W^2+H^2)$.

Compared to the vanilla transformer, whose space complexity takes up to $\mathcal{O}(H^2W^2)$, our Stripformer is more token-efficient, which only takes $\mathcal{O}(HW(H+W)+H^2+W^2) = \mathcal{O}(HW(H+W))$. Furthermore, the proposed horizontal and vertical multi-head Intra-SA and Inter-SA can help explore blur orientations. Stacking interlaced Intra-SA and Inter-SA blocks can reveal blur magnitudes. Therefore, even though Stripformer is a transformer-based architecture, our meticulous design for deblurring not only demands less memory but also achieves superior deblurring performance. 

\subsection{Loss Function}\label{subsec:loss}
\noindent\textbf{Contrastive Learning.}
Contrastive learning~\cite{chen2020simple} is known to be an effective self-supervised technique. It allows a model to generate universal features from data similarity and dissimilarity even without labels. Recently, it has been adopted in vision tasks~\cite{IPT,contrastive_dehaze} by pulling close ``positive" (similar) pairs and pushing apart ``negative" (dissimilar) pairs in the feature space. 
Motivated by~\cite{contrastive_dehaze}, we utilize contrastive learning to make a deblurred output image similar to its ground truth but dissimilar to its blurred input. Let the blurred input be $X$, its deblurred result be $R$, and the associated sharp ground truth be $S$, where $X$, $R$, and $S \in R^{H \times W \times 3}$. We regard $X$, $R$, and $S$ as the negative, anchor, and positive samples.
The contrastive loss is formulated as
\begin{equation}
    L_{con} = \frac{L{_1}\big(\psi(S) - \psi(R)\big)}{L{_1}\big(\psi(X) - \psi(R)\big)},
    %L_{con} = \frac{\mathbf{SumAbs}\big((\psi(S) - \psi(R))\odot M\big)}{\mathbf{SumAbs}\big((\psi(X) - \psi(R))\odot M\big)},
\end{equation}
where $\psi$ extracts the hidden features 
from conv3-2 of the fixed pre-trained VGG-19~\cite{vgg19}, and $L{_1}$ represents the L${_1}$ norm. Minimizing $L_{con}$ helps pull the deblurred result $R$ close to the sharp ground truth $S$ (the numerator) while pushing $R$ away from its blurred input $X$ (the denominator) in the same latent feature space.

\noindent\textbf{Optimization.}
The loss function of Stripformer for deblurring is 
\begin{equation}
    L = L_{char} + \lambda_{1} L_{edge} + \lambda_{2} L_{con},
\end{equation}
where $ L_{char}$ and $L_{edge}$ are the Charbonnier loss and the edge loss the same as those used in MPRNet~\cite{Zamir2021MPRNet}, and $L_{con}$ is the  contrastive loss. Here, we set to $\lambda_{1}=0.05$ as set in~\cite{Zamir2021MPRNet} and $\lambda_{2}=0.0005$.

\section{Experiments}
In this section, the proposed Stripformer is evaluated. 
We first describe the datasets and implementation details. Then we compare our method with the state-of-the-arts quantitatively and qualitatively. 
At last, the ablation studies are provided to demonstrate the effectiveness of the Stripformer design. 
\subsection{Datasets and Implementation Details}
For comparison, we adopt the widely used GoPro dataset \cite{Nah_2017_CVPR} which includes $2,103$ blurred and sharp pairs for training and $1,111$ pairs for testing. 
The HIDE dataset~\cite{HAdeblur} with $2,025$ images is included only for testing. 
To address real-world blurs, we evaluate our method on the RealBlur dataset \cite{rim_2020_ECCV}, which has $3,758$ blurred and sharp pairs for training and $980$ pairs for testing.\footnoteonlytext{The authors from the universities in Taiwan completed the experiments.}

We train our network with a batch size of $8$ on the GoPro dataset. 
Adam optimizer is used with the initial learning rate of $10^{-4}$ that is steadily decayed to $10^{-7}$ by the cosine annealing strategy. 
We adopt random cropping, flipping, and rotation for data augmentation, like~\cite{MT_2020_ECCV,Yuan_2020_CVPR}. We train Stripformer on the GoPro training set and evaluate it on the GoPro testing set and HIDE dataset. For the RealBlur dataset, we use the RealBlur training set to train the model and evaluate it on the RealBlur testing set. We test our method on the full-size images using an NVIDIA 3090 GPU. 

\begin{table}[t!]
\centering
\setlength{\tabcolsep}{3mm}
\caption{Evaluation results on the benchmark GoPro testing set. The best two scores in each column are highlighted in bold and underlined, respectively. $\dag$ represents the work did not release the code or pre-trained weight. Params and Time are calculated in (M) and (ms), respectively.}
\begin{tabular}{l|c|c|c|c}\hline\hline
{Method}         & {PSNR\,$\uparrow$}  & {SSIM\,$\uparrow$} & {Params\,$\downarrow$} &{Time\,$\downarrow$} \\ \hline
\multicolumn{4}{c}{CNN-based} 
\\\hline
{DeblurGAN-v2~\cite{Kupyn_2019_ICCV}} & {29.55} & {0.934} & {68} & {60}\\
{EDSD\dag~\cite{Yuan_2020_CVPR}} & {29.81} & {0.937} & \bf{3} & \bf{10}\\
{SRN~\cite{tao2018srndeblur}} & {30.25} & {0.934} & \underline{7} & {650}  \\
{PyNAS\dag~\cite{Hu_2021_ICCV}} & {30.62} & {0.941} & 9 & \underline{17}  \\
{DSD~\cite{gao2019dynamic}} & {30.96} & {0.942} & {\bf{3}} & {1300}\\
{DBGAN\dag~\cite{Zhang_2020_CVPR}} & {31.10} & {0.942} & {-} & {-} \\
{MTRNN~\cite{MT_2020_ECCV}} & {31.13} & {0.944} & {\bf3} & {30} \\
{DMPHN~\cite{Zhang_2019_CVPR}} & {31.20} & {0.945} & {22} & 303\\
{SimpleNet\dag~\cite{Li_2021_ICCV}} & {31.52} & {0.950} & {25} & 376\\
{RADN\dag~\cite{RADN_2020_ECCV}} & {31.85} & {0.953} & {-} & 38 \\
{SAPHN\dag~\cite{SAPN2020}} & {32.02} & {0.953} & {-} & 770 \\
{SPAIR\dag~\cite{Purohit_2021_ICCV}} & {32.06} & {0.953} & {-} & - \\
{MIMO~\cite{MIMO}} & {32.45} & {0.957} & {16} & 31  \\
{TTFA\dag~\cite{Chi_2021_CVPR}} & {32.50} & {0.958} & {-} & -  \\
{MPRNet~\cite{Zamir2021MPRNet}} & {32.66} & {0.959} & {20} & 148\\ \hline
\multicolumn{4}{c}{Transformer-based}
\\\hline
{IPT$^\dag$~\cite{IPT}} & {32.58} & {-} & {114} & {-}      \\
{Stripformer} & {\bf{33.08}} & {\bf{0.962}} & {20} & 52  \\
\hline\hline
\end{tabular}
\label{Tab:GoPro_eval}
\end{table}

\subsection{Experimental Results}
\noindent\textbf{Quantitative Analysis.}
We compare our model on the GoPro testing set with several existing SOTA methods~\cite{IPT,Chi_2021_CVPR,MIMO,gao2019dynamic,Hu_2021_ICCV,Kupyn_2019_ICCV,Li_2021_ICCV,MT_2020_ECCV,RADN_2020_ECCV,Purohit_2021_ICCV,SAPN2020,tao2018srndeblur,Yuan_2020_CVPR,Zamir2021MPRNet,Zhang_2019_CVPR,Zhang_2020_CVPR}.
In Table~\ref{Tab:GoPro_eval}, all of the compared methods utilize CNN-based architectures to build the deblurring networks except for IPT~\cite{IPT} and our network, where transformers serve as the backbone for deblurring.
As shown in Table~\ref{Tab:GoPro_eval}, Stripformer performs favorably against all competing methods in both PSNR and SSIM on the GoPro test set.
It is worth mentioning that Stripformer can achieve state-of-the-art performance by only using the GoPro training set. It has exceeded the expectation that transformer-based architectures tend to have suboptimal performance compared to most of the CNN-based methods without using a large amount of training data~\cite{IPT,dosovitskiy2020vit}.
That is, a transformer-based model typically requires a large dataset, \eg more than one million annotated data, for pre-training to compete with a CNN-based model in vision tasks.
For example, IPT fine-tunes the models with pre-training on ImageNet for deraining, denoising, and super-resolution tasks to achieve competitive performance.
We attribute Stripformer's success on deblurring to being able to better leverage the local and global information with our intra-strip and inter-strip attention design.
In addition, Stripformer can run efficiently without using recurrent architectures compared to the~\cite{SAPN2020,Zamir2021MPRNet,Zhang_2019_CVPR}.
Table~\ref{Tab:HIDE_eval} and Table~\ref{Tab:Realblur_eval} report more results on the HIDE and RealBlur datasets, respectively. As can be seen, Stripformer again achieves the best deblurring performance among the SOTA methods for both synthetic and real-world blur datasets.

\begin{table}[t!]
\centering
\setlength{\tabcolsep}{3mm}
\caption{Evaluation results on the benchmark HIDE dataset. Note that all the models are trained on the GoPro training set. The best two scores in each column are highlighted in bold and underlined, respectively. Params and Time are calculated in (M) and (ms), respectively.}
\begin{tabular}{l|c|c|c|c}\hline\hline
{Method} & {PSNR\,$\uparrow$}  & {SSIM\,$\uparrow$} & {Params\,$\downarrow$} &{Time\,$\downarrow$} 
\\\hline
{DeblurGAN-v2~\cite{Kupyn_2019_ICCV}}  & {27.40} & {0.882} & 68 &  57  \\
{SRN~\cite{tao2018srndeblur}}        & {28.36} & {0.904} & \underline7 &  424      
\\
{HAdeblur~\cite{HAdeblur}}        & {28.87} & {0.930} & - &  -  
\\
{DSD~\cite{gao2019dynamic}}        & {29.01} & {0.913}  & \bf3 &  1200  
\\
{DMPHN~\cite{Zhang_2019_CVPR}}        & {29.10} & {0.918}  & 22 & 310  
\\
{MTRNN~\cite{MT_2020_ECCV}}        & {29.15} & {0.918} & 22 & \underline{40}  
\\
{SAPHN$^\dag$~\cite{SAPN2020}}        & {29.98} & {0.930} & - &  -
\\
{MIMO~\cite{MIMO}}  & {30.00} & {0.930} & 16 &  \bf30  
\\ 
{TTFA$^\dag$~\cite{Chi_2021_CVPR}} & {30.55} & {0.935}  & - &  -
\\
{MPRNet~\cite{Zamir2021MPRNet}}        & {\underline{30.96}} & {\underline{0.939}}
& 20 & 140 
\\
\noalign{\hrule height 1.0pt}
{Stripformer}      & \bf{31.03} & \bf{0.940}   
& 20 &  43
\\
\hline\hline
\end{tabular}
\label{Tab:HIDE_eval}
\end{table}

\begin{figure}[t!]
\centering
\includegraphics[width=1\columnwidth]{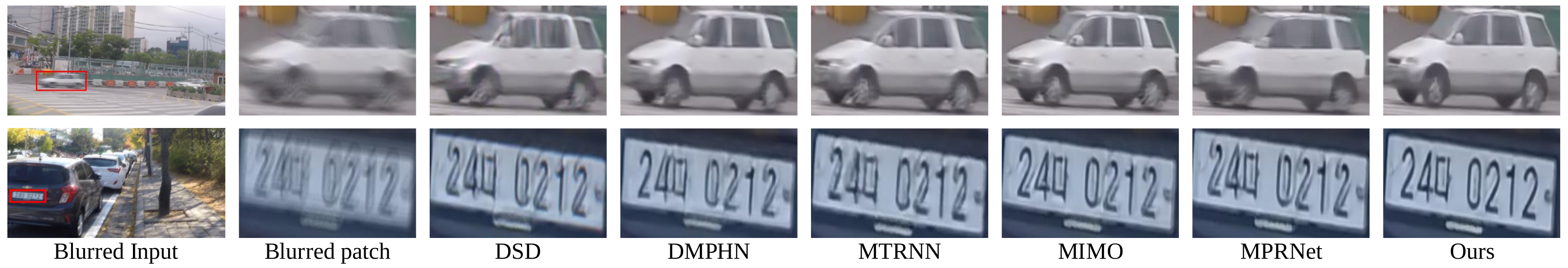}
\caption{Qualitative comparisons on the GoPro testing set. The deblurred results from left to right are produced by DSD~\cite{gao2019dynamic}, DMPHN~\cite{Zhang_2019_CVPR}, MTRNN~\cite{MT_2020_ECCV}, MIMO~\cite{MIMO}, MPRNet~\cite{Zamir2021MPRNet}, and our method, respectively.}
\label{fig:Visualization_Result_GoPro}
\end{figure}

\begin{figure}[t!]
\centering
\includegraphics[width=1\columnwidth]{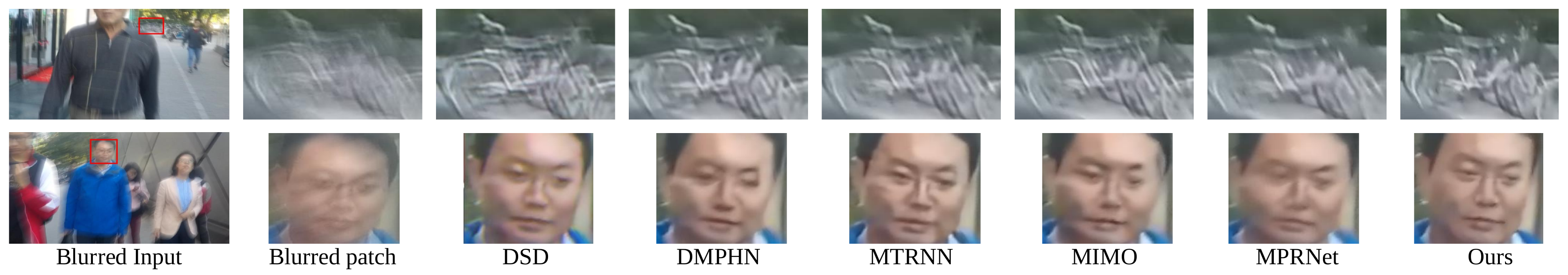}
\caption{Qualitative comparisons on the HIDE dataset. The deblurred results from left to right are produced by DSD~\cite{gao2019dynamic}, DMPHN~\cite{Zhang_2019_CVPR}, MTRNN~\cite{MT_2020_ECCV}, MIMO~\cite{MIMO}, MPRNet~\cite{Zamir2021MPRNet} and our method, respectively.}
\label{fig:Visualization_Result_HIDE}
\end{figure}

\begin{figure*}[t!]
\centering
\includegraphics[width=1\columnwidth]{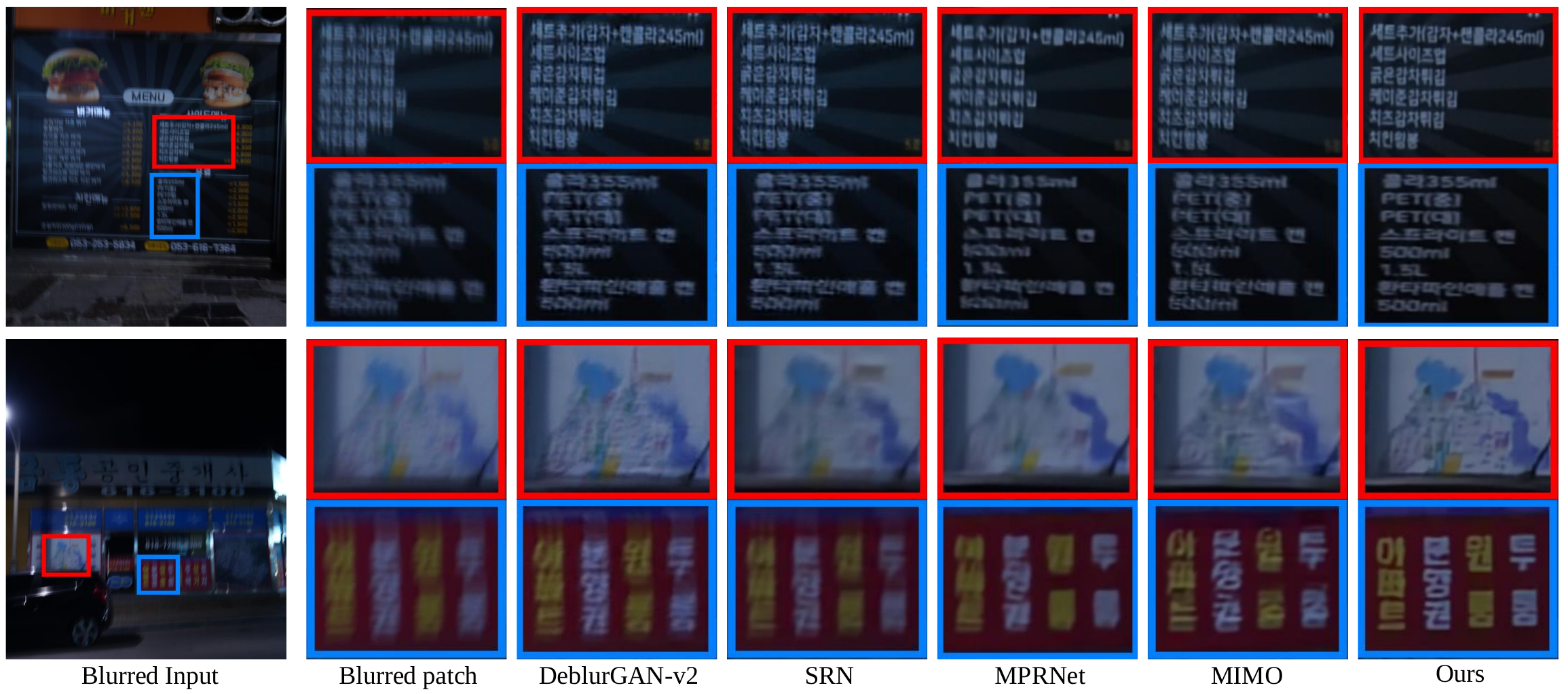}
\caption{Qualitative comparisons on the RealBlur dataset. The deblurred results from left to right are produced by DeblurGAN-v2~\cite{Kupyn_2019_ICCV}, SRN~\cite{tao2018srndeblur}, MPRNet~\cite{Zamir2021MPRNet}, MIMO~\cite{MIMO} and our method.}
\label{fig:Visualization_result_Realblur}
\end{figure*}

\begin{figure*}[t!]
\centering
\includegraphics[width=1\columnwidth]{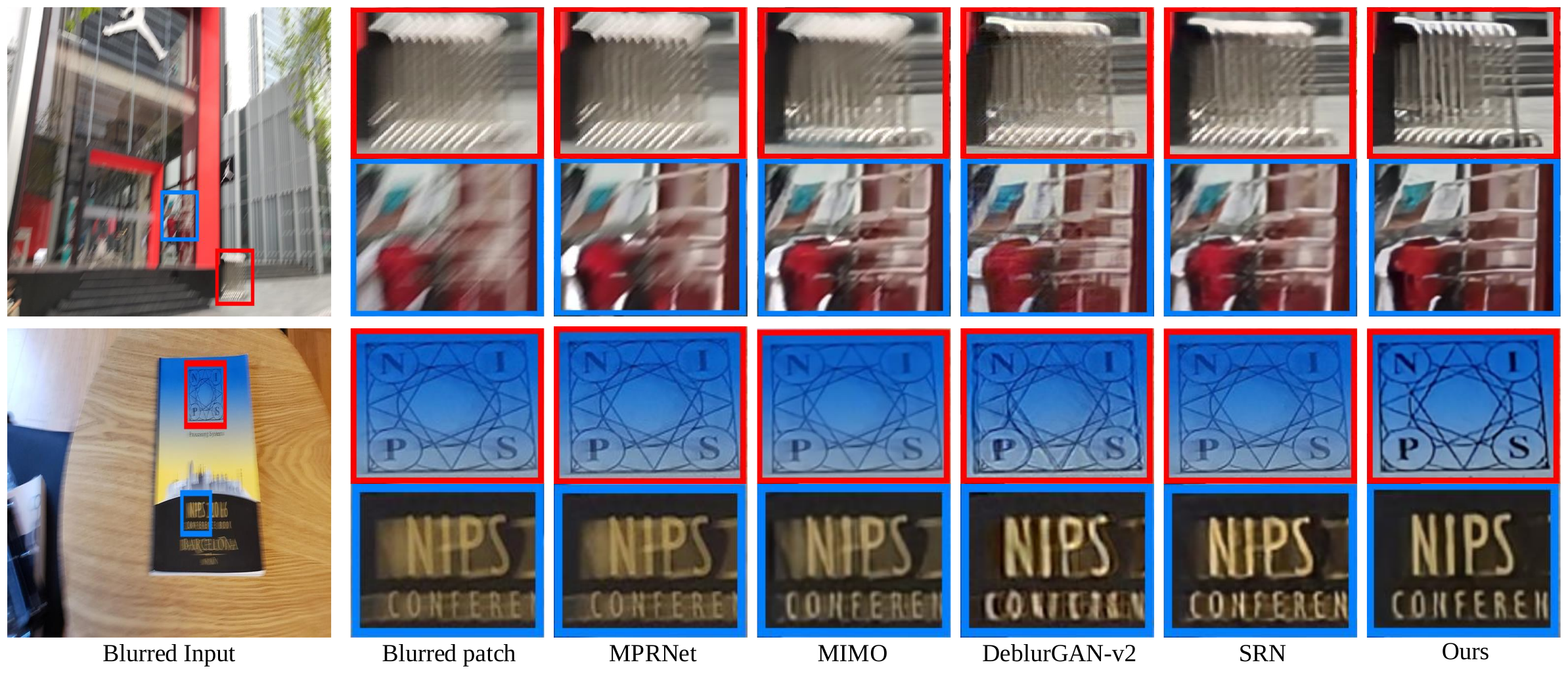}
\caption{Qualitative comparisons on the RWBI~\cite{Zhang_2020_CVPR}. The deblurred results from left to right are produced by MPRNet~\cite{Zamir2021MPRNet}, MIMO~\cite{MIMO}, DeblurGAN-v2~\cite{Kupyn_2019_ICCV}, SRN~\cite{tao2018srndeblur}, and our method.}
\label{fig:Visualization_result_RWBI}
\end{figure*}

\begin{table}[t!]
\centering
\setlength{\tabcolsep}{0.5mm}
\caption{Evaluation results on the RealBlur testing set. The best and the second scores in each column are highlighted in bold and underlined, respectively. Params and Time are calculated in (M) and (ms), respectively.
}
\begin{tabular}{l|cc|cc|cc}
\hline\hline
            & \multicolumn{2}{c|}{RealBlur-J} & \multicolumn{2}{c|}{RealBlur-R} & \multicolumn{2}{c}{RealBlur}\\
Model      & PSNR\,$\uparrow$ & SSIM\,$\uparrow$ & PSNR\,$\uparrow$ & SSIM\,$\uparrow$ & {Params\,$\downarrow$} &{Time\,$\downarrow$}  \\ \hline
DeblurGANv2~\cite{Kupyn_2019_ICCV}  & 29.69          & 0.870  & 36.44  & 0.935 & 68 & 60         \\ 
SRN~\cite{tao2018srndeblur} & 31.38  & 0.909          & 38.65  & 0.965 & \bf7 & 412 \\          
MPRNet~\cite{Zamir2021MPRNet}  & 31.76          &  \underline{0.922}          & \underline{39.31}           & \underline{0.972}  & 20 & 113       \\
{SPAIR$^\dag$~\cite{Purohit_2021_ICCV}} & {31.82} & {-} & {-} & - & -  & - \\
MIMO~\cite{MIMO}  & \underline{31.92}   & 0.919          & -          & -    & \underline{16} & \bf39      \\ 
\noalign{\hrule height 1.0pt}
Stripformer        &    \bf32.48      &   \bf0.929       &     \bf39.84       &     \bf0.974 &  20 & \underline{42}    \\
\hline\hline
\end{tabular}
\label{Tab:Realblur_eval}
\end{table}
\noindent\textbf{Qualitative Analysis.}
Figure~\ref{fig:Visualization_Result_GoPro} and Figure~\ref{fig:Visualization_Result_HIDE} show the qualitative comparisons on the GoPro test set and the HIDE dataset among our method and those in \cite{MIMO,gao2019dynamic,MT_2020_ECCV,Zamir2021MPRNet,Zhang_2019_CVPR}.
They indicate that our method can better restore images, especially on highly textured regions such as texts and vehicles.
It can also restore fine-grained information like facial expressions on the HIDE dataset. 
In Figure~\ref{fig:Visualization_result_Realblur}, we show the qualitative comparisons on the RealBlur test set among our method and those in \cite{MIMO,Kupyn_2019_ICCV,tao2018srndeblur,Zamir2021MPRNet}. 
This dataset contains images in low-light environments where motion blurs usually occur. 
As can be observed, our model can better restore these regions than the competing works.
In Figure~\ref{fig:Visualization_result_RWBI}, we show the qualitative comparisons on the RWBI~\cite{Zhang_2020_CVPR} dataset, which contains real images without ground truth. As shown, our model produces sharper deblurring results than the other methods~\cite{MIMO,Kupyn_2019_ICCV,tao2018srndeblur,Zamir2021MPRNet}.
Overall, the qualitative results demonstrate that Stripformer works well on blurred images in both synthetic and real-world scenes.

\begin{table}[t!]
\centering
\setlength{\tabcolsep}{4.5mm}
\caption{Component analysis of Stripformer trained and tested on the GoPro training and test sets.}
\begin{tabular}{cccc|c}
\hline
Intra-SA & Inter-SA & CPE & $L_{con}$ & PSNR \\
\hline
$\surd$ &       &     &       & 32.84 \\
      & $\surd$ &     &       & 32.88 \\
$\surd$ & $\surd$ &     &       & 33.00 \\
$\surd$ & $\surd$ & $\surd$ &       & 33.03 \\
$\surd$ & $\surd$ & $\surd$ & $\surd$ & 33.08 \\
\hline
\end{tabular}
\label{tab:ablation}
\end{table}

\subsection{Ablation Studies}
%We conduct ablation studies to analyze the proposed design.
%
Here, We conduct ablation studies to analyze the proposed design, including component and computational analyses and comparisons against various modern attention mechanisms.
%
%At last, we evaluate the effect of pre-training with a larger dataset.

\noindent\textbf{Component Analysis.} 
Stripformer utilizes intra-strip and inter-strip attention blocks in horizontal and vertical directions to address various blur patterns with diverse magnitudes and orientations.
Table~\ref{tab:ablation} reports the contributions of individual components of Stripformer.
The first two rows of Table~\ref{tab:ablation} show the performance of using either the intra-strip or inter-strip attention blocks only, respectively. 
The two types of attention blocks are synergistic since combining them results in better performance, as given in the third row.
It reveals that Stripformer encoders feature both pixel-wise and region-wise dependency, more suitable to solve the regional-specific blurred artifacts. 
In the fourth row, the conditional positional encoding (CPE)~\cite{chu2021conditional} is included for positional encoding and boosts the performance, which shows it works better for arbitrary input sizes compared to the fixed, learnable positional encoding.  
The last row shows that the contrastive loss can further improve deblurring performance.

\begin{table}[t]
\centering
\setlength{\tabcolsep}{1mm}
\caption{Comparison among various attention mechanisms. FLOPs are calculated the same way as IPT. The inference time (Time) is measured based on the GoPro and HIDE datasets on average for full-resolution (1280x720) images.}
\begin{tabular}{c|ccccc}
\hline
Method &  IPT~\cite{IPT} & Swin~\cite{liu2021Swin} & CCNet~\cite{huang2019ccnet} & Twins~\cite{chu2021Twins} & Ours \\
\hline
Params (M) & 114 &  20   &   20    &   20   &   20   \\
FLOPs (G) &  32 &  6.7   &   7.4  &   6.5   &    6.9  \\
Time (ms) & -- &   48  &   47    &   43   &   48   \\
GoPro (PSNR) & 32.58 &  32.39   &  32.73 &  32.89  &  \bf33.08\\
HIDE (PSNR) & -- &  30.19  & 30.61  & 30.82   &  \bf31.03\\
\hline
\end{tabular}
\label{tab:transformer}
\end{table}

\noindent\textbf{Analysis on Transformer-based Architectures and Attention Mechanisms.}
We compare Stripformer against efficient transformer-based architectures, including Swin~\cite{liu2021Swin} and Twins~\cite{chu2021Twins}, and the attention mechanism CCNet~\cite{huang2019ccnet}. 
Note that these three compared methods are designed for efficient attention computation rather than deblurring.
For fair comparisons, we replace our attention mechanism in Stripformer with theirs using a similar parameter size and apply the resultant models to deblurring.
As reported in Table~\ref{tab:transformer}, the proposed intra-strip and inter-strip attention mechanism in Stripformer works better in PSNR than all the other competing attention mechanisms. 
The reason is that Stripformer takes the inductive bias of image deblurring into account to design the intra-strip and inter-strip tokens and attention mechanism.
It strikes a good balance among the image restorability, model size, and computational efficiency for image deblurring.
Besides, SWin works with a shifted windowing scheme, restricting its self-attention computed locally. Thus, it is not enough to obtain sufficient global information for deblurring like our intra- and inter-strip attention design. CCNet extracts pixel correlations horizontally and vertically in a criss-cross manner. Twins uses local window attention and global sub-sampled attention (down to the size of $7\times7$). Both of them consider more global information, working better than SWin. Our strip-wise attention design can better harvest local and global blur information to remove short-range and long-range blur artifacts, performing favorably against all these compared attention mechanisms.

\section{Conclusions}  
In this paper, we propose a novel model, called Stripformer, for dynamic scene image deblurring. 
Stripformer is a token- and parameter-efficient transformer-based model designed for region-specific blur artifacts in images taken in dynamic scenes. To better address such blur having diverse orientations and magnitudes, Stripformer utilizes both intra- and inter-strip attentions to demand not only less memory and computation costs than a vanilla transformer but to achieve excellent deblurring performance.
Experimental results show that our method achieves SOTA performance on three benchmarks, including the GoPro, HIDE, and Realblur datasets, without using a large dataset like ImageNet for pre-training. 
Moreover, in terms of memory usage, model size, and inference time, Stripformer performs quite competitively.
We believe that Stripformer is a friendly transformer-based model that can serve as a good basis for further advancing transformer-based architectures in image deblurring and more.
\section{Acknowledgments}
This work was supported in part by the Ministry of Science and Technology (MOST) under grants 109-2221-E-009-113- MY3, 111-2628-E-A49-025-MY3, 111-2634-F-007-002, 110-2634-F-002-050, 110-2634-F-006-022, 110-2622-E-004-001, and 111-2221-E-004-010. This work was funded in part by Qualcomm through a Taiwan University Research Collaboration Project and by MediaTek. We thank the National Center for High-performance Computing (NCHC) of National Applied Research Laboratories (NARLabs) in Taiwan for providing computational and storage resources.

\clearpage
\bibliographystyle{splncs04}
\bibliography{egbib}

\begin{thebibliography}{10}
\providecommand{\url}[1]{\texttt{#1}}
\providecommand{\urlprefix}{URL }
\providecommand{\doi}[1]{https://doi.org/#1}

\bibitem{DETR}
Carion, N., Massa, F., Synnaeve, G., Usunier, N., Kirillov, A., Zagoruyko, S.:
  End-to-end object detection with transformers. In: Proc. Euro. Conf. Computer
  Vision (2020)

\bibitem{IPT}
Chen, H., Wang, Y., Guo, T., Xu, C., Deng, Y., Liu, Z., Ma, S., Xu, C., Xu, C.,
  Gao, W.: Pre-trained image processing transformer. In: Proc. Conf. Computer
  Vision and Pattern Recognition (2021)

\bibitem{chen2020simple}
Chen, T., Kornblith, S., Norouzi, M., Hinton, G.: A simple framework for
  contrastive learning of visual representations. In: Proc. Int'l Conf. Machine
  Learning (2020)

\bibitem{Chi_2021_CVPR}
Chi, Z., Wang, Y., Yu, Y., Tang, J.: Test-time fast adaptation for dynamic
  scene deblurring via meta-auxiliary learning. In: Proc. Conf. Computer Vision
  and Pattern Recognition (2021)

\bibitem{MIMO}
Cho, S.J., Ji, S.W., Hong, J.P., Jung, S.W., Ko, S.J.: Rethinking
  coarse-to-fine approach in single image deblurring. In: Proc. Int'l Conf.
  Computer Vision (2021)

\bibitem{Cho_2009_ACM}
Cho, S., Lee, S.: Fast motion deblurring. In: ACM Trans. Graphic. (2009)

\bibitem{chu2021Twins}
Chu, X., Tian, Z., Wang, Y., Zhang, B., Ren, H., Wei, X., Xia, H., Shen, C.:
  Twins: Revisiting the design of spatial attention in vision transformers. In:
  Proc. Neural Information Processing Systems (2021)

\bibitem{chu2021conditional}
Chu, X., Tian, Z., Zhang, B., Wang, X., Wei, X., Xia, H., Shen, C.: Conditional
  positional encodings for vision transformers. In: arxiv preprint
  arXiv:2102.10882 (2021)

\bibitem{dosovitskiy2020vit}
Dosovitskiy, A., Beyer, L., Kolesnikov, A., Weissenborn, D., Zhai, X.,
  Unterthiner, T., Dehghani, M., Minderer, M., Heigold, G., Gelly, S.,
  Uszkoreit, J., Houlsby, N.: An image is worth 16x16 words: Transformers for
  image recognition at scale. In: Proc. Int'l Conf. Learning Representations
  (2021)

\bibitem{Fergus2006}
Fergus, R., Singh, B., Hertzmann, A., Roweis, S.T., Freeman, W.T.: Removing
  camera shake from a single photograph. In: ACM Trans. Graphic. (2006)

\bibitem{gao2019dynamic}
Gao, H., Tao, X., Shen, X., Jia, J.: Dynamic scene deblurring with parameter
  selective sharing and nested skip connections. In: Proc. Conf. Computer
  Vision and Pattern Recognition (2019)

\bibitem{hou2020strip}
Hou, Q., Zhang, L., Cheng, M.M., Feng, J.: {Strip Pooling}: Rethinking spatial
  pooling for scene parsing. In: Proc. Conf. Computer Vision and Pattern
  Recognition (2020)

\bibitem{hu2018senet}
Hu, J., Shen, L., Sun, G.: Squeeze-and-excitation networks. In: Proc. Conf.
  Computer Vision and Pattern Recognition (2018)

\bibitem{Hu_2021_ICCV}
Hu, X., Ren, W., Yu, K., Zhang, K., Cao, X., Liu, W., Menze, B.: Pyramid
  architecture search for real-time image deblurring. In: Proc. Int'l Conf.
  Computer Vision (2021)

\bibitem{huang2019ccnet}
Huang, Z., Wang, X., Huang, L., Huang, C., Wei, Y., Liu, W.: Ccnet: Criss-cross
  attention for semantic segmentation. In: Proc. Int'l Conf. Computer Vision
  (2019)

\bibitem{5206802}
{Joshi}, N., {Zitnick}, C.L., {Szeliski}, R., {Kriegman}, D.J.: Image
  deblurring and denoising using color priors. In: Proc. Conf. Computer Vision
  and Pattern Recognition (2009)

\bibitem{Kupyn_2019_ICCV}
Kupyn, O., Martyniuk, T., Wu, J., Wang, Z.: Deblurgan-v2: Deblurring
  (orders-of-magnitude) faster and better. In: Proc. Int'l Conf. Computer
  Vision (2019)

\bibitem{Li_2021_ICCV}
Li, J., Tan, W., Yan, B.: Perceptual variousness motion deblurring with light
  global context refinement. In: Proc. Int'l Conf. Computer Vision (2021)

\bibitem{liu2021Swin}
Liu, Z., Lin, Y., Cao, Y., Hu, H., Wei, Y., Zhang, Z., Lin, S., Guo, B.: Swin
  transformer: Hierarchical vision transformer using shifted windows. In: Proc.
  Int'l Conf. Computer Vision (2021)

\bibitem{Nah_2017_CVPR}
Nah, S., Kim, T.H., Lee, K.M.: Deep multi-scale convolutional neural network
  for dynamic scene deblurring. In: Proc. Conf. Computer Vision and Pattern
  Recognition (2017)

\bibitem{6909767}
Pan, J., Hu, Z., Su, Z., Yang, M.H.: Deblurring text images via l0-regularized
  intensity and gradient prior. In: Proc. Conf. Computer Vision and Pattern
  Recognition (2014)

\bibitem{Pan_2016_cvpr}
Pan, J., Sun, D., Pfister, H., Yang, M.H.: Deblurring images via dark channel
  prior. In: Proc. Conf. Computer Vision and Pattern Recognition (2018)

\bibitem{MT_2020_ECCV}
{Park}, D., {Kang}, D.U., {Kim}, J., {Chun}, S.Y.: Multi-temporal recurrent
  neural networks for progressive non-uniform single image deblurring with
  incremental temporal training. In: Proc. Euro. Conf. Computer Vision (2020)

\bibitem{parmar2018image}
Parmar, N., Vaswani, A., Uszkoreit, J., Kaiser, {\L}., Shazeer, N., Ku, A.,
  Tran, D.: Image transformer. arXiv preprint arXiv:1802.05751  (2018)

\bibitem{RADN_2020_ECCV}
{Purohit}, K., {Rajagopalan}, A.N.: Region-adaptive dense network for efficient
  motion deblurring. In: Proc. Nat'l Conf. Artificial Intelligence (2020)

\bibitem{Purohit_2021_ICCV}
Purohit, K., Suin, M., Rajagopalan, A.N., Boddeti, V.N.: Spatially-adaptive
  image restoration using distortion-guided networks. In: Proc. Int'l Conf.
  Computer Vision (2021)

\bibitem{Ranftl21}
Ranftl, R., Bochkovskiy, A., Koltun, V.: Vision transformers for dense
  prediction. In: Proc. Int'l Conf. Computer Vision (2021)

\bibitem{rim_2020_ECCV}
Rim, J., Lee, H., Won, J., Cho, S.: Real-world blur dataset for learning and
  benchmarking deblurring algorithms. In: Proc. Euro. Conf. Computer Vision
  (2020)

\bibitem{HAdeblur}
Shen, Z., Wang, W., Shen, J., Ling, H., Xu, T., Shao, L.: Human-aware motion
  deblurring. In: Proc. Int'l Conf. Computer Vision (2019)

\bibitem{vgg19}
Simonyan, K., Zisserman, A.: Very deep convolutional networks for large-scale
  image recognition. In: Proc. Int'l Conf. Learning Representations (2015)

\bibitem{SAPN2020}
{Suin}, M., {Purohit}, K., Rajagopalan, A.N.: Spatially-attentive
  patch-hierarchical network for adaptive motion deblurring. In: Proc. Conf.
  Computer Vision and Pattern Recognition (2020)

\bibitem{Jian_2015_CVPR}
Sun, J., Cao, W., Xu, Z., Ponce, J.: Learning a convolutional neural network
  for non-uniform motion blur removal. In: Proc. Conf. Computer Vision and
  Pattern Recognition (2015)

\bibitem{tao2018srndeblur}
Tao, X., Gao, H., Shen, X., Wang, J., Jia, J.: Scale-recurrent network for deep
  image deblurring. In: Proc. Conf. Computer Vision and Pattern Recognition
  (2018)

\bibitem{vaswani2017attention}
Vaswani, A., Shazeer, N., Parmar, N., Uszkoreit, J., Jones, L., Gomez, A.N.,
  Kaiser, {\L}., Polosukhin, I.: Attention is all you need. In: Proc. Neural
  Information Processing Systems (2017)

\bibitem{wang2018non}
Wang, X., Girshick, R., Gupta, A., He, K.: Non-local neural networks. In: Proc.
  Conf. Computer Vision and Pattern Recognition (2018)

\bibitem{contrastive_dehaze}
Wu, H., Qu, Y., Lin, S., Zhou, J., Qiao, R., Zhang, Z., Xie, Y., Ma, L.:
  Contrastive learning for compact single image dehazing. In: Proc. Conf.
  Computer Vision and Pattern Recognition (2021)

\bibitem{yang2020learning}
Yang, F., Yang, H., Fu, J., Lu, H., Guo, B.: Learning texture transformer
  network for image super-resolution. In: Proc. Conf. Computer Vision and
  Pattern Recognition (2020)

\bibitem{Yuan_2020_CVPR}
Yuan, Y., Su, W., Ma, D.: Efficient dynamic scene deblurring using spatially
  variant deconvolution network with optical flow guided training. In: Proc.
  Conf. Computer Vision and Pattern Recognition (2020)

\bibitem{Zamir2021MPRNet}
Zamir, S.W., Arora, A., Khan, S., Hayat, M., Khan, F.S., Yang, M.H., Shao, L.:
  Multi-stage progressive image restoration. In: Proc. Conf. Computer Vision
  and Pattern Recognition (2021)

\bibitem{yan2020sttn}
Zeng, Y., Fu, J., Chao, H.: Learning joint spatial-temporal transformations for
  video inpainting. In: Proc. Euro. Conf. Computer Vision (2020)

\bibitem{SAGAN_2019_PMLR}
Zhang, H., Goodfellow, I., Metaxas, D., Odena, A.: Self-attention generative
  adversarial networks. In: arXiv preprint arXiv:1805.08318 (2019)

\bibitem{Zhang_2019_CVPR}
Zhang, H., Dai, Y., Li, H., Koniusz, P.: Deep stacked hierarchical multi-patch
  network for image deblurring. In: Proc. Conf. Computer Vision and Pattern
  Recognition (2019)

\bibitem{Zhang_2020_CVPR}
Zhang, K., Luo, W., Zhong, Y., Ma, L., Stenger, B., Liu, W., Li, H.: Deblurring
  by realistic blurring. In: Proc. Conf. Computer Vision and Pattern
  Recognition (2020)

\end{thebibliography}
\end{document}